\documentclass[10pt,twocolumn,letterpaper]{article}

\usepackage{wacv}
\usepackage{times}
\usepackage{epsfig}
\usepackage{graphicx}
\usepackage{amsmath}
\usepackage{amssymb}

\def\wacvPaperID{107} 

\wacvfinalcopy 

\ifwacvfinal
\def\assignedStartPage{9876} 
\fi
\ifwacvfinal
\usepackage[breaklinks=true,bookmarks=false]{hyperref}
\else
\usepackage[pagebackref=true,breaklinks=true,colorlinks,bookmarks=false]{hyperref}
\fi

\ifwacvfinal
\else
\fi

\begin{document}

\title{Improving Few-Shot Learning using Composite Rotation based Auxiliary Task}


\author{
Pratik Mazumder$^{\dagger}$ \hspace{2cm}Pravendra Singh$^{\dagger}$ \hspace{2cm}Vinay P. Namboodiri$^{\dagger \ast}$\\
$^{\dagger}$Department of Computer Science and Engineering, IIT Kanpur, India\\
$^{\ast}$University of Bath, United Kingdom \\
{\tt\small \{pratikm, psingh\}@iitk.ac.in, vpn22@bath.ac.uk}
}

\maketitle

\begin{abstract}
In this paper, we propose an approach to improve few-shot classification performance using a composite rotation based auxiliary task. Few-shot classification methods aim to produce neural networks that perform well for classes with a large number of training samples and classes with less number of training samples. They employ techniques to enable the network to produce highly discriminative features that are also very generic. Generally, the better the quality and generic-nature of the features produced by the network, the better is the performance of the network on few-shot learning. Our approach aims to train networks to produce such features by using a self-supervised auxiliary task. Our proposed composite rotation based auxiliary task performs rotation at two levels, i.e., rotation of patches inside the image (inner rotation) and rotation of the whole image (outer rotation) and assigns one out of 16 rotation classes to the modified image. We then simultaneously train for the composite rotation prediction task along with the original classification task, which forces the network to learn high-quality generic features that help improve the few-shot classification performance. We experimentally show that our approach performs better than existing few-shot learning methods on multiple benchmark datasets. 

\end{abstract}

\vspace{-10pt}
\section{Introduction}

Deep learning techniques have been used extensively to tackle several computer vision tasks, and they have been very successful \cite{russakovsky2015imagenet,krizhevsky2012imagenet,zhou2014learning}. The ability of neural networks to learn informative features from images is the main factor behind the successes of deep learning frameworks. However, neural networks need to be trained on large volumes of labeled data, which is a cause of concern since obtaining labeled data can be very difficult and sometimes very expensive. Obtaining labeled data may also require manual annotation, which is time-consuming and costly. In many real-world cases, it is not feasible to collect a large amount of labeled data for all categories of data. In such a case, the network will not perform well for classes with few labeled training examples. On the other hand, humans can learn new categories of images from very few samples and recognize them in the wild with high probability. Deep learning models generally do not possess such a capability and are hence not human-like. Researchers have been looking into ways to achieve this. Few-shot learning is a step in this direction.

In the few-shot learning setting, the networks are trained in such a way that they can perform well for classes with few training examples and classes with many training examples. This can be achieved when the network has the ability to extract highly discriminative features from input images. The network should be trained in such a way that it can extract discriminative features even for a new set of categories. Few-shot learning methods generally operate on episodes. Episodes are tiny-datasets with a small train set and a small test set. Each episode consists of examples from a fixed small number of classes.

There have been many works in few-shot learning. Prototypical network \cite{snell2017prototypical} computes class prototypes and then uses the nearest neighbor-based classification to predict classes for query images. MAML \cite{finn2017model} trains the network to quickly adapt to a new set of classes to perform classification on them. LEO \cite{rusu2018metalearning} learns to generate weights for classifier using the support examples of the classes in the episode. RFS \cite{tian2020rethinking} proposes to improve the quality of representation produced by the network by using knowledge distillation.

\begin{figure*}[t]
    \centering
     \includegraphics[width=\textwidth]{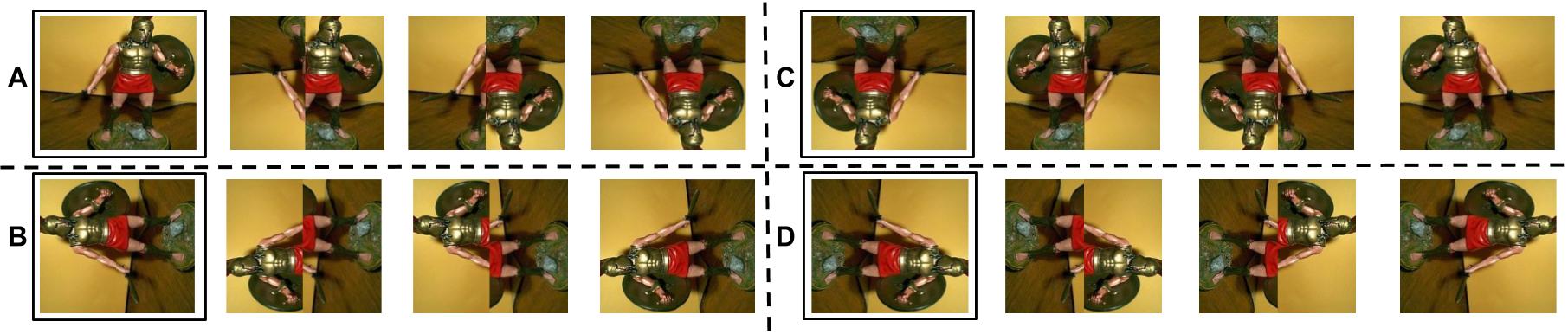}
    \caption{\textbf{Composite rotation classes created from a single image.} The four images, which are inside a separate box, represent the 4 outer rotations used by  \cite{gidaris2018unsupervised}. Our composite rotation first rotates the entire image by 0, 90, 180, or 270 degrees (A,B,C,D) (outer rotation), then splits the image vertically from the middle and rotates each half by 0 or 180 degrees (inner rotation). This creates 16 combinations, which become the 16 classes of our self-supervision method.}
    \label{fig:method}
    \vspace{-15pt}
\end{figure*}

Since the performance of few-shot learning methods heavily depends on the discriminative and generic nature of the features extracted by its network, researchers have been looking to improve networks and enable them to extract such features. One standard method of improving the representation/features produced by a network is to use self-supervised learning. Self-supervised learning techniques do not require labeled data to train. Self-supervised learning is a type of unsupervised learning which involves creating artificial labels using unlabeled examples. Artificial labels are created in a way that is simple, fast, and automated, using only the visual data contained in the input. Training on such pseudo-labeled examples provides a pseudo supervised learning environment that helps the network to gain the ability to extract good and generic features from images. Self-supervised learning trains networks by making use of the structural information contained in the input images. Some widely used self-supervision techniques include training the network to predict the angle of rotation applied to the input image \cite{gidaris2018unsupervised}, to predict the relative position of a patch of an image with respect to a given patch of the image \cite{doersch2015unsupervised,noroozi2016unsupervised}, to produce color images from gray-scale images \cite{larsson2016learning,zhang2016colorful}. 

Since self-supervised learning improves the features produced by a network and does not require additional labeled data, it is a great candidate to be used to improve few-shot learning. We propose to train the few-shot network on an auxiliary self-supervised task based on composite rotation. Our proposed composite rotation divides the image into patches and rotates them inside the image in addition to rotating the whole image (see Fig. \ref{fig:method}). The network is trained to predict the type of composite rotation that has been applied to the image along with predicting the actual label. This multi-task training forces the model to learn more detailed features of the objects contained inside the images.

In order to validate the efficacy of our auxiliary task, we incorporate it into the method (RFS) proposed in \cite{tian2020rethinking}. RFS makes use of knowledge distillation to improve the representational quality of the network, thereby improving its few-shot classification performance. We show experimentally that our auxiliary task can further significantly improve RFS.  A simple rotation based auxiliary task is proposed in \cite{gidaris2019boosting}. However, the simple rotation operation rotates the entire image as a whole (outer rotation). Therefore, the relative position between the points in the image does not change. This can be considered as a bias in the data. Due to this bias in the data, the features produced by the network trained using this technique can still be improved to capture more meaningful details about the objects in the image. This can be done by rotating patches inside the image (inner rotation) in addition to the outer rotation. Our proposed composite rotation performs both inner and outer rotations. For a fair comparison with \cite{gidaris2019boosting}, we replace their rotation based auxiliary task with our composite rotation based auxiliary task (CRAT) and perform few-shot learning experiments. We experimentally show that our proposed auxiliary task performs significantly better than the simple rotation based auxiliary task used in \cite{gidaris2019boosting}. Our approach is described in detail in Sec \ref{sec:method}. 

The contributions of this paper are as follows:
\begin{itemize}
    \item We propose an approach to improve few-shot classification that trains the network on a composite rotation based auxiliary self-supervised task along with the general classification task.
    
    \item We incorporate our proposed auxiliary task into multiple few-shot learning methods \cite{tian2020rethinking, gidaris2019boosting} and experimentally show that we are able to improve significantly over them.
    
    \item We perform experiments on multiple benchmark few-shot learning datasets to show the efficacy of our approach. We also perform ablation experiments to validate our approach.
\end{itemize}

\section{Related Works}
\subsection{Few-Shot Learning}
 Many few-shot learning methods have been proposed by researchers \cite{finn2017model,snell2017prototypical,Flennerhag2020Meta-Learning,gidaris2018dynamic,sung2018learning,chen2019diversity,tian2020rethinking,simon2020adaptive,guo2020attentive,li2020adversarial}. Prototypical Network \cite{snell2017prototypical} first averages the embeddings of the support data-points of each class produced by a base network to obtain a ``prototype" for each class and then finds the class embedding closest to the embedding of the query image. Model Agnostic Meta Learning \cite{finn2017model} optimizes the network in such a way that it can quickly adapt to a new episode. A graph neural network architecture is used in \cite{garcia2018fewshot}. Meta Network \cite{munkhdalai2017meta} performs fast parameterization of the underlying network for rapid generalization.
 
The method in \cite{ren2018metalearning} applies the prototypical network to a semi-supervised setup by making use of labeled and unlabeled examples in each episode. TADAM \cite{oreshkin2018tadam} uses a task-based embedding as an attention to the convolutional layers of the base network. This attention helps shift the image embeddings closer to the embeddings of similar images based on the classes that are being trained on. RelationNet \cite{sung2018learning} uses relation scores to match query and support images of the classes. Learning without forgetting \cite{gidaris2018dynamic} uses support examples of the novel classes and classifier weights of the base classes to learn classifier weights for the novel classes. R2D2 \cite{bertinetto2018metalearning} makes use of fast convergent methods like ridge regression for few-shot learning. LEO \cite{rusu2018metalearning} solves an optimization problem in the parameter space to learn good parameters for the few-shot classifier. 

MetaOptNet \cite{lee2019meta} learns more discriminative features by making use of linear predictors.  TPN~\cite{liu2019learning} uses a graph-based method to exploit the test data itself to improve few-shot classification in a transductive setting. In \cite{simon2020adaptive}, the authors propose to use dynamic classifiers constructed from limited samples. The method proposed in \cite{guo2020attentive} improves the generation of classifier weights for few-shot classification by maximizing the mutual information between the weights and the data. In \cite{li2020adversarial}, the authors use conditional Wasserstein GAN to hallucinate highly discriminative features. 

RFS \cite{tian2020rethinking} makes use of self-distillation, which is a knowledge transfer process from a trained network to a student network with the same architecture. The authors show that this results in the student network learning better and more generic features. Such features enable the student network to be more generalizable and perform better on few-shot learning. Recently, self-supervision has been applied to few-shot learning. In \cite{gidaris2019boosting}, the authors use self-supervised techniques as auxiliary tasks to improve few-shot learning. The proposed method trains the model simultaneously on an auxiliary self-supervised task in addition to its original task. The images are modified to create self-supervised tasks such as rotation or relative patch position prediction.  The authors show that using such auxiliary tasks helps the network perform better in the few-shot learning settings, indicating that such a network extracts a more generic set of features from the images.

Our paper focuses on training the network using a composite rotation based auxiliary task (CRAT) along with the classification task. Our composite rotation is more sophisticated than the simple rotation used in \cite{gidaris2019boosting} and performs better than it for all the benchmark datasets that we experiment on. We also use our technique to improve RFS \cite{tian2020rethinking} and achieve state-of-the-art results.

\subsection{Self-Supervised Learning Techniques}

Many self-supervision techniques have been proposed to improve semantic feature learning. In \cite{pathak2016context}, the authors train the network to perform image inpainting or image completion. Several works have proposed to use the variance of image colorization as self-supervision \cite{larsson2016learning,zhang2016colorful}. The method proposed in \cite{doersch2015unsupervised} trains the network to predict the relative position of a patch of an image with respect to a given patch of the same image. A convolutional neural network is trained to solve Jigsaw puzzles in \cite{noroozi2016unsupervised}. 

In \cite{gidaris2018unsupervised}, the authors rotate images by a fixed set of angles and then train the network to predict the angle of rotation. The method proposed in \cite{dosovitskiy2014discriminative} forms surrogate classes by transforming images in specific ways and trains convolutional neural networks to predict the class of the image.  In \cite{feng2019self}, the authors propose to train the network to learn additional discriminative features along with the rotation angle prediction. This is achieved by additionally training the network to reduce the distance between different versions of the same image, which have been rotated at a different angle. This ensures that the network learns to produce features that have a better instance-level discriminative ability.
  
Contrastive Multiview Coding (CMC) \cite{tian2019contrastive} takes different views of an image and trains the network to learn a representation that maximizes the mutual information between the different views. But CMC requires specialized architecture, including separate encoders for different views of the data. Momentum Contrast (MoCo) \cite{he2020momentum} performs matching of encoded queries q to a dictionary of encoded keys using a contrastive loss in order to train the network. But MoCo requires a memory bank to store the dictionary. SimCLR \cite{chen2020simple} applies separate sets of data augmentation to the input resulting in two different but correlated views and uses contrastive loss to bring them closer in the feature space. It does not require specialized architectures or a memory bank and still achieves state-of-the-art unsupervised learning results, outperforming CMC and MoCo. We compare an auxiliary task based on SimCLR with our proposed composite rotation based auxiliary task in Sec. \ref{sec:ablation_ss}.

\vspace{-4pt}

\section{Method}\label{sec:method}
\subsection{Problem Setting}
In the few-shot learning setting, the train and test classes are referred to as base and novel classes, respectively. Here, the networks operate on episodes of data. An episode can be thought of as a small dataset that is further divided into a mini-train set and a mini-test set. Each episode consists of a fixed small number of classes $N$, and each class has $K$ support training examples. Such episodes are known as $N$-way $K$-shot episodes. The test set consists of query data points belonging to one of the $N$ classes.

\begin{figure*}[t]
    \centering
    \includegraphics[width=\textwidth]{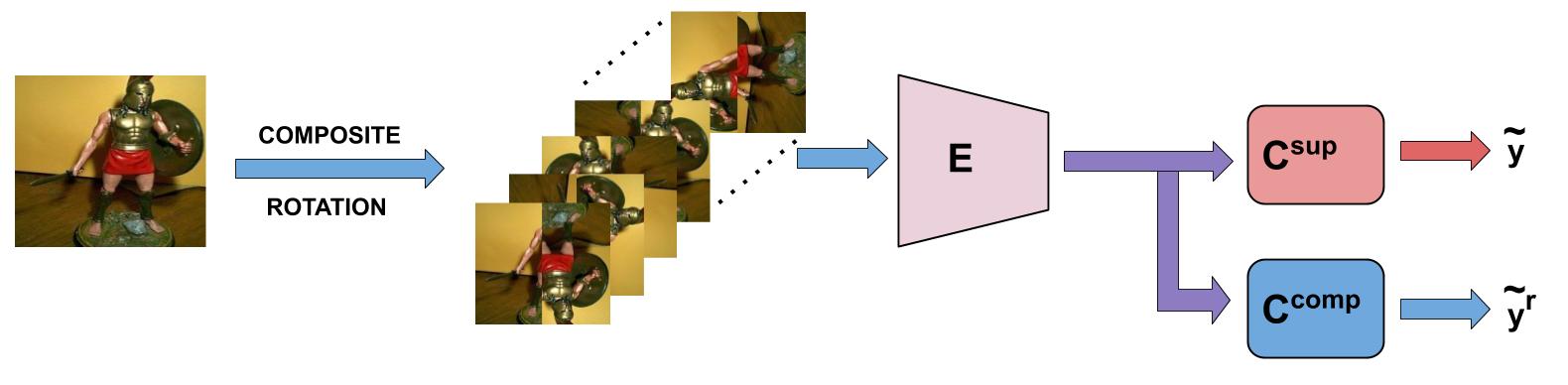}
    \caption{Self-supervised auxiliary classification task. The input image is rotated using our composite rotation technique. The rotated images are fed to the feature extraction unit $E$. $C^{sup}$ and $C^{comp}$ perform fully-supervised classification and auxiliary composite rotation classification on the features extracted by $E$.}
\label{fig:ss}
    \vspace{-15pt}
\end{figure*}
\subsection{Proposed Composite Rotation based Auxiliary Task (CRAT)}
We propose a composite rotation based auxiliary task (CRAT) that helps in improving few-shot learning performance. In our proposed composite rotation, we first rotate the entire image by either 0, 90, 180, or 270 degrees. Next, we split the image vertically from the middle. Then, we rotate each half by either 0 or 180 degrees (see Fig. \ref{fig:method}). The outer rotation angle remains the same for a total of 4 combinations, while the inner rotation on each half is either 0 or 180 degrees. Therefore, there are a total of 16 classes of composite rotation. We validate the operations used in this composite rotation through ablation experiments in Sec \ref{sec:ablation}. We use this composite rotation as an auxiliary task to improve few-shot classification methods. A few-shot learning network generally consists of a feature extraction network $E$, that extracts features from images. We add a  composite rotation classification network $C^{comp}$ for our proposed auxiliary task. For any given input image, we take the output of the feature extraction network $E$ and feed it to $C^{comp}$ to predict the composite rotation class. The training on this auxiliary task is carried out in parallel to the main classification task. After the training process is completed, the composite rotation classification network $C^{comp}$ is discarded as it has served its purpose of improving the feature extraction network $E$. The composite rotation based auxiliary task loss $L_{CR}$ can be defined as 
\begin{equation}\label{eq:ss1}
    L_{CR}(E,C^{comp}) = \Sigma_i f_{CE}(\widetilde{y_i^{r}},y_i^{r})
\end{equation}
where $i$ refers to the $i^{th}$ data point in the mini-batch, $f_{CE}$ refers to the cross-entropy loss function, $\widetilde{y_i^{r}}$ refers to the composite rotation class of the $i^{th}$ data point predicted by $C^{comp}$, and $y_i^{r}$ refers to the actual composite rotation class of the $i^{th}$ data point.

Our composite rotation technique will cause some objects within the image to behave differently after rotation since the rotation is not being applied only to the entire image as a whole. There will be cases where semantic parts of the same object in the image will get rotated differently. Objects may also end up rotating to different centers of rotation. Therefore, the relative position of objects in the image and even the relative position of the semantic parts of the same object may get changed after our composite rotation. This will force the network to extract more detailed features about the objects, their parts, and their orientations. As a result, networks trained on these types of pseudo-classes should be better at extracting meaningful features from images.

Ideally, we should create even more subdivisions within the image and rotate them in order to complicate the task even more. However, this will exponentially increase the total number of classes and, hence, the model's complexity, which might make it difficult to train. We tried to split the image into four quarters and rotate each of them through the four angles apart from the full image rotation, which led to a total of 1024 classes. The resulting auxiliary task is very difficult, and the network performance is not adequate on few-shot classification, as shown in Sec. \ref{sec:ablation}.

\subsection{Integrating CRAT with RFS \cite{tian2020rethinking}}
In \cite{tian2020rethinking}, the authors propose a few-shot learning method (RFS) focused on improving the feature representation quality of the network. The model consists of a feature extraction network $E$, a fully-supervised classification network $C^{sup}$, and a linear few-shot classification module $C^{few}$. The method proposed in \cite{tian2020rethinking} first trains the feature extraction network $E$ and the fully-supervised classification network $C^{sup}$ on the entire training set (phase 1). The trained network from phase 1 is used as a teacher ($E_T, C^{sup}_T$) to train a student network ($E_S, C^{sup}_S$) with the same architecture using self-distillation multiple times (phase 2). For self-distillation, the student is trained to minimize the Kullback-Leibler divergence (KL) between the soft predictions of the teacher and the student networks. During phase 2, the student is trained on the fully supervised classification loss on the entire training set and on the self-distillation loss. $C^{sup}_S$ is discarded after the training process. During testing, for every testing episode, $E_S$ is used to extract features for the support examples of each class. These features are used to train a logistic regression classification model $C^{few}_S$. Finally, $E_S$ is used to extract the query sample features, and $C^{few}_S$ is used to classify them. 

We introduce our composite rotation based auxiliary task (CRAT) into this method to improve it. A composite rotation classification network $C^{comp}$ is added to the model. During training, $E$ is used to extract features for the rotated image. This feature is fed to both $C^{sup}$ and $C^{comp}$ to perform fully supervised classification and composite rotation classification, respectively (see Fig. \ref{fig:ss}). This will help the feature extractor learn to extract better features in order to perform better for both the tasks. 

In this modified training procedure, the phase 1 now involves training the feature extraction network $E$, the fully-supervised classification network $C^{sup}$ and the composite rotation classification network $C^{comp}$ on the entire training set. In phase 2, the trained network from phase 1 is used as a teacher ($E_T, C^{sup}_T, C^{comp}_T$) to train a student network ($E_S, C^{sup}_S, C^{comp}_S$) with the same architecture using the fully-supervised classification loss, the composite rotation prediction loss, and the self-distillation loss. During testing, both $C^{sup}_S$ and $C^{comp}_S$ are discarded. $E_S$ extracts features for the support examples of each class, which are used to train a  logistic regression classification model $C^{few}_S$. Finally, $C^{few}_S$ is used to classify the query samples. We experimentally show that our proposed composite rotation based auxiliary task can significantly improve the performance of RFS. The new phase 1 and 2 losses can be defined as:
\begin{equation}\label{eq:rfs_phase1}
 L_{phase1} = \Sigma_i (f_{CE}(\widetilde{y_i},y_i) + \lambda f_{CE}(\widetilde{y_i^{r}},y_i^{r})) 
\end{equation}
where, $y_i$ is the real label for the $i^{th}$ data point, $\widetilde{y_i}$ refers to the label predicted by $C^{sup}$  for the $i^{th}$ data point, $\widetilde{y_i^{r}}$ refers to the composite rotation class predicted by $C^{comp}$  for the $i^{th}$ data point, $y_i^{r}$ refers to the actual composite rotation class  of the $i^{th}$ data point, $\lambda$ hyper-parameter determines the influence of the auxiliary task loss in the total loss.
\begin{multline}\label{eq:rfs_phase2}
 L_{phase2} = \Sigma_i [\alpha(f_{CE}(\widetilde{y_i},y_i) + \lambda f_{CE}(\widetilde{y_i^{r}},y_i^{r})) \\+  \beta KL(\phi_S(x_i), \phi_T(x_i)) ]
\end{multline}
where, KL refers to the KL-Divergence function, $\alpha,\beta$ are hyper-parameters, $\phi_S(x_i),\phi_T(x_i)$ refer to the soft label predictions of the student and teacher models for the $i^{th}$ data point. $\phi_S(x_i)=F_{softmax}(C^{sup}_S(E_S(x_i)))$, $\phi_T(x_i)=F_{softmax}(C^{sup}_T(E_T(x_i)))$ where $F_{softmax}$ is the softmax function.

\subsection{Integrating CRAT with BF3S \cite{gidaris2019boosting}}
The method (BF3S) proposed in \cite{gidaris2019boosting} also uses an auxiliary task. It uses a simple rotation \cite{gidaris2018unsupervised} based self-supervised auxiliary task to train the feature extraction network simultaneously along with the few-shot classification task. 
After the training step, the auxiliary task network is discarded, and the remaining network is used to perform few-shot classification. 
The simple rotation based self-supervised task rotates the image by a fixed angle (out of $0^{o},90^{o},180^{o},270^{o}$), and then trains the network to predict the angle of rotation. This does not change the relative position of points inside the image, and all points rotate with the same center of rotation, i.e., the center of the image. This leads to a bias in the data. We replace this simple rotation based auxiliary task with our composite rotation based auxiliary task and perform the same training process as proposed in BF3S. We experimentally show that our composite rotation based auxiliary task performs significantly better than the auxiliary tasks originally used in this method.

\section{Experiments}
\subsection{Datasets}
We perform few-shot classification experiments on 4 benchmark datasets: mini-ImageNet \cite{vinyals2016matching}, tiered-ImageNet \cite{ren2018metalearning}, CIFAR-FS \cite{bertinetto2018metalearning} and FC-100 \cite{oreshkin2018tadam}. 

mini-ImageNet \cite{vinyals2016matching} consists of 100 classes, each of which has around 600 images of size $84\times 84$ pixels. The classes are divided into 64 train classes, 16 validation classes, and 20 test classes. It has been derived from the ImageNet \cite{russakovsky2015imagenet} dataset. tiered-ImageNet \cite{ren2018metalearning} is also derived from ImageNet and consists of 351 train, 97 validation, and 60 test classes. CIFAR-FS and Few-shot-CIFAR100 (FC-100) are both derived from CIFAR-100 \cite{krizhevsky2009learning}  dataset. CIFAR-FS consists of 64 train, 16 validation, and 20 test classes with images of size $32\times32$ pixels. FC-100 consists of 60 train, 20 validation, and 20 test classes with images of size $32\times32$ pixels. FC-100 splits classes based on the super-classes in CIFAR-100 in order to minimize similarity between classes from different splits. 

\subsection{Implementation Details}
In order to show the efficacy of our proposed auxiliary task, we experiment with RFS \cite{tian2020rethinking} and BF3S \cite{gidaris2019boosting}. We modify RFS \cite{tian2020rethinking} to include our proposed composite rotation based auxiliary task (CRAT), and we replace the auxiliary task in BF3S \cite{gidaris2019boosting} with CRAT. For the experiments involving RFS \cite{tian2020rethinking}, we use the ResNet-12 architecture for the feature extraction network $E$. It consists of 4 residual blocks, each having 3 convolutional layers with kernel size $3\times3$. A max-pooling layer of kernel size $2\times2$ is added after each of the initial 3 residual blocks, and a global average pooling layer is added after the last residual block. The composite rotation classifier is implemented as a convolutional neural network with 4 convolutional blocks. Each block consists of a convolutional layer of 640 convolutional filters of kernel size of $3\times3$, a batch normalization layer, and a ReLU activation function. Each convolutional block has 640 output filters. An adaptive average pooling block is added after the last convolutional block, and it is followed by a fully-connected layer with input size 640 and output size 16. The $\lambda$ hyper-parameter, that decides the contribution of our auxiliary task loss to the total loss is taken as 1 for all experiments. We take $\alpha=0.5$ and $\beta=1$ for the RFS experiments. During training, we create multiple copies of the same image with different composite rotation transformations. To keep the process scalable, we apply only five transformations per image at a time while classifying for all the 16 transformations.  The other settings used for these experiments are the same as given in \cite{tian2020rethinking}.

For the experiments involving BF3S \cite{gidaris2019boosting},
we conduct experiments on the best performing model of \cite{gidaris2019boosting} that uses the WideResNet-28-10 \cite{zagoruyko2016wide} architecture as the feature extractor and the Cosine Classifier (CC). It is a 28-layer Wide Residual Network \cite{zagoruyko2016wide} with width factor 10, which produces a feature map of size $10\times10\times640$ and a global average pooling converts this to a 640-dimensional feature vector. For the composite-rotation based auxiliary task classification network, we use a 4-residual-layer residual block followed by a global average pooling block (similar to \cite{gidaris2019boosting}) and a fully connected classification layer with an output size of 16. For the BF3S experiments, the settings are the same as given in \cite{gidaris2019boosting}. The classification performance is estimated using an average of the classification accuracy over the test images achieved by the network in each task/episode. We present the results for the 5-way 1-shot and 5-shot settings.

\subsection{mini-ImageNet Results}
\begin{table}[!t]

\caption{Average 1/5-shot 5-way few-shot classification accuracy over test images from the novel classes of the mini-ImageNet dataset. $^\ast$  indicate methods that train on a union of train and validation (train+val) set.}

\scalebox{0.75}{
\addtolength{\tabcolsep}{-4.3pt}
\begin{tabular}{|l|l|c|c|}
\hline
\multicolumn{1}{|l|}{Models} & Backbone & \multicolumn{1}{|c|}{1-shot} & \multicolumn{1}{c|}{5-shot}\\
\hline
\hline
MAML~\cite{finn2017model}  (ICML'17)                    & Conv-4-64  & 48.70 $\pm$ 1.84\% & 63.11 $\pm$ 0.92\%\\
Proto Net~\cite{snell2017prototypical} (NIPS'17) & Conv-4-64  & 49.42 $\pm$ 0.78\% & 68.20 $\pm$ 0.66\%\\
MetaNet\cite{munkhdalai2017meta}(ICML'17)    & ResNet-12      & 57.10 $\pm$ 0.70\% & 70.04 $\pm$ 0.63\%\\
LwoF~\cite{gidaris2018dynamic}  (CVPR'18)               & Conv-4-64  & 56.20 $\pm$ 0.86\% & 72.81 $\pm$ 0.62\%\\
RelationNet~\cite{sung2018learning} (CVPR'18)           & Conv-4-64  & 50.44 $\pm$ 0.82\% & 65.32 $\pm$ 0.70\%\\
GNN~\cite{garcia2018fewshot}  (ICLR'18)                     & Conv-4-64  & 50.30\%          & 66.40\%\\
SNAIL~\cite{mishra2018a}    (ICLR'18)             & ResNet-12      & 55.71 $\pm$ 0.99\% & 68.88 $\pm$ 0.92\%\\
Qiao et al. \cite{qiao2018few}$^\ast$     (CVPR'18)           & WRN-28-10 & 59.60 $\pm$ 0.41\% & 73.74 $\pm$ 0.19\%\\
TADAM~\cite{oreshkin2018tadam} (NIPS'18)                & ResNet-12      & 58.50 $\pm$ 0.30\% & 76.70 $\pm$ 0.30\%\\
TPN~\cite{liu2019learning}   (ICLR'19)                 & Conv-4-64 & 55.51 $\pm$ 0.86\% & 69.86 $\pm$ 0.65\%\\
R2-D2~\cite{bertinetto2018metalearning} (ICLR'19)          & Conv-4-64  & 49.50 $\pm$ 0.20\% & 65.40 $\pm$ 0.20\%\\
R2-D2~\cite{bertinetto2018metalearning} (ICLR'19)               & Conv-4-512 & 51.80 $\pm$ 0.20\% & 68.40 $\pm$ 0.20\%\\
STANet~\cite{Yan2019ADA} (AAAI'19) & ResNet-12 & 58.35 $\pm$ 0.57\% & 71.07 $\pm$ 0.39\%\\
IdeMe-Net~\cite{Chen_2019} (CVPR'19) & ResNet-18 & 59.14 $\pm$ 0.86\% & 74.63 $\pm$0.74\% \\
Shot-Free~\cite{Ravichandran_2019} (ICCV'19) & ResNet-12 & 59.04\% & 77.64\%\\
SalNet Intra\cite{Zhang_2019}(CVPR'19)& ResNet-101 & 62.22 $\pm$ 0.87\% & 77.95 $\pm$ 0.65\%\\
LEO$^\ast$~\cite{rusu2018metalearning}  (ICLR'19)                   & WRN-28-10 & 61.76 $\pm$ 0.08\% & 77.59 $\pm$ 0.12\%\\
BF3S CC+Rot\cite{gidaris2019boosting}(ICCV'19)                      & WRN-28-10  & 62.93 $\pm$ 0.45\% & 79.87 $\pm$ 0.33\%\\
MetaOptNet\cite{lee2019meta}(CVPR'19)                & ResNet-12      & 62.64 $\pm$ 0.61\% & 78.63 $\pm$ 0.46\%\\
MetaOptNet$^\ast$\cite{lee2019meta}(CVPR'19)                & ResNet-12      & 64.09 $\pm$ 0.62\% & 80.00 $\pm$ 0.45\%\\
RFS\cite{tian2020rethinking} & ResNet-12  & 64.82 $\pm$ 0.60\% & 82.14 $\pm$ 0.43\%\\
RFS\cite{tian2020rethinking}$^\ast$ & ResNet-12  & 66.58 $\pm$ 0.65\% & 83.22 $\pm$ 0.39\%\\
Warp-MAML~\cite{Flennerhag2020Meta-Learning}(ICLR'20) & Conv-4-64 & 52.30 $\pm$ 0.80\% & 68.40 $\pm$ 0.60\%\\
D-SVS\cite{chen2019variational}(AAAI'20) & ResNet-12 & 60.16 $\pm$ 0.47\% & 77.25 $\pm$ 0.15\%\\
Deep DTN~\cite{chen2019diversity}  (AAAI'20)                   & ResNet-12 & 63.45 $\pm$ 0.86\% & 77.91$\pm$ 0.62\%\\
AFHN~\cite{li2020adversarial} (CVPR'20) & ResNet-18 & 62.38 $\pm$ 0.72\% & 78.16 $\pm$ 0.56\%\\
AWGIM\cite{guo2020attentive} (CVPR'20) & WRN-28-10  & 63.12 $\pm$ 0.08\% & 78.40 $\pm$ 0.11\%\\
DSN-MR\cite{simon2020adaptive} (CVPR'20) & ResNet-12  & 64.60 $\pm$ 0.72\% & 79.51 $\pm$ 0.50\%\\
DSN-MR\cite{simon2020adaptive}$^\ast$ (CVPR'20) & ResNet-12  & 67.09 $\pm$ 0.68\% & 81.65 $\pm$ 0.69\%\\
\hline
\hline
BF3S CC\cite{gidaris2019boosting}+CRAT (Ours) & WRN-28-10  & 63.87 $\pm$ 0.47\% & 80.92 $\pm$ 0.34\%\\
\textbf{RFS\cite{tian2020rethinking} + CRAT} \textbf{(Ours)} & ResNet-12  & \textbf{68.44 $\pm$ 0.60\%} & \textbf{83.75 $\pm$ 0.41\%}\\
\textbf{RFS\cite{tian2020rethinking} + CRAT}$^\ast$ \textbf{(Ours)}  & ResNet-12  & \textbf{68.89 $\pm$ 0.61\%} & \textbf{84.86 $\pm$ 0.45\%}\\
\hline
\end{tabular}}
\label{tab:imagenet}
\vspace{-7pt}
\end{table}

The results for few-shot classification on the mini-ImageNet dataset are given in Table \ref{tab:imagenet}. The results indicate that using CRAT significantly improves both BF3S \cite{gidaris2019boosting} and RFS \cite{tian2020rethinking}. In the case of BF3S, CC + CRAT outperforms CC + Rot \cite{gidaris2019boosting} by absolute margins of 0.94\% and 1.05\% for the 1-shot 5-way and 5-shot 5-way settings, respectively. RFS + CRAT performs better than RFS by absolute margins of 3.62\% and 1.61\% for the 1-shot 5-way and 5-shot 5-way settings, respectively. RFS + CRAT$^\ast$, which is trained on the combined training and validation set, performs better than RFS$^\ast$ (train+val) by absolute margins of 2.31\% and 1.64\% for the 1-shot 5-way and 5-shot 5-way settings respectively. RFS + CRAT performs better than existing state-of-the-art methods for both train and train+val settings.

\subsection{tiered-ImageNet Results}
\begin{table}[!t]
\centering
\caption{Average 1/5-shot 5-way few-shot classification accuracy over test images from the novel classes of the tiered-ImageNet dataset. $^\ast$  indicate methods that train on a union of train and validation (train+val) set.}
\scalebox{0.75}{
\addtolength{\tabcolsep}{-4.5pt}
\begin{tabular}{|l|l|c|c|}
\hline
\multicolumn{1}{|l|}{Models} & Backbone & \multicolumn{1}{|c|}{1-shot} & \multicolumn{1}{c|}{5-shot}\\
\hline
\hline
MAML~\cite{finn2017model} (ICML'17)                  & Conv-4-64  & 51.67 $\pm$ 1.81\% & 70.30 $\pm$ 0.08\%\\
Proto Net~\cite{snell2017prototypical}  (NIPS'17) & Conv-4-64  & 53.31 $\pm$ 0.89\% & 72.69 $\pm$ 0.74\%\\
RelationNet~\cite{sung2018learning}   (CVPR'18)         & Conv-4-64  & 54.48 $\pm$ 0.93\% & 71.32 $\pm$ 0.78\%\\
TPN~\cite{liu2019learning}   (ICLR'19)                 & Conv-4-64 & 59.91 $\pm$ 0.94\% & 73.30 $\pm$ 0.75\%\\
LEO$^\ast$ ~\cite{rusu2018metalearning}      (ICLR'19)               & WRN-28-10 & 66.33 $\pm$ 0.05\% & 81.44 $\pm$ 0.09\%\\
MetaOptNet~\cite{lee2019meta}(CVPR'19)                & ResNet-12      & 65.99 $\pm$ 0.72\% & 81.56 $\pm$ 0.53\%\\
MetaOptNet$^\ast$~\cite{lee2019meta}(CVPR'19)                & ResNet-12      & 65.81 $\pm$ 0.74\% & 81.75 $\pm$ 0.53\%\\
Shot-Free~\cite{Ravichandran_2019} (ICCV'19) & ResNet-12 & 66.87\% & 82.64\%\\
BF3S CC+Rot \cite{gidaris2019boosting}(ICCV'19)                         & WRN-28-10  & 70.53 $\pm$ 0.51\% & 84.98 $\pm$ 0.36\%\\
RFS\cite{tian2020rethinking} & ResNet-12  & 71.52 $\pm$ 0.69\% & 86.03 $\pm$ 0.49\%\\
RFS\cite{tian2020rethinking}$^\ast$ & ResNet-12  & 72.98 $\pm$ 0.71\% & 87.46 $\pm$ 0.44\%\\
Warp-MAML\cite{Flennerhag2020Meta-Learning} (ICLR'20) & Conv-4-64 & 57.20 $\pm$ 0.90\% & 74.10 $\pm$ 0.70\%\\
AWGIM\cite{guo2020attentive} (CVPR'20) & WRN-28-10  & 67.69 $\pm$ 0.11\% & 82.82 $\pm$ 0.13\%\\
DSN-MR\cite{simon2020adaptive} (CVPR'20) & ResNet-12  & 67.39 $\pm$ 0.82\% & 82.85 $\pm$ 0.56\%\\
DSN-MR\cite{simon2020adaptive}$^\ast$ (CVPR'20) & ResNet-12  & 68.44 $\pm$ 0.77\% & 83.32 $\pm$ 0.66\%\\
\hline
\hline
BF3S CC\cite{gidaris2019boosting}+CRAT(Ours) & WRN-28-10  & 71.37 $\pm$ 0.45\% & 85.89 $\pm$ 0.36\%\\
\textbf{RFS\cite{tian2020rethinking} + CRAT} \textbf{(Ours)} & ResNet-12  & \textbf{73.45 $\pm$ 0.83\%} & \textbf{87.33 $\pm$ 0.47\%}\\
\textbf{RFS\cite{tian2020rethinking} + CRAT}$^\ast$ \textbf{(Ours)}  & ResNet-12  & \textbf{74.63 $\pm$ 0.78\%} & \textbf{88.67 $\pm$ 0.44\%}\\
\hline
\end{tabular}
}
\label{tab:tiered}
\vspace{-15pt}
\end{table}

Table \ref{tab:tiered} reports the results for few-shot classification on the tiered-ImageNet dataset. BF3S with our proposed auxiliary task (CC + CRAT) performs better than CC + Rot \cite{gidaris2019boosting}. RFS + CRAT performs significantly better than RFS \cite{tian2020rethinking}. Our RFS + CRAT outperforms existing state-of-the-art methods for both train and train+val settings as shown in Table \ref{tab:tiered}.

\subsection{CIFAR-FS Results}
\begin{table}[t]
\centering
\caption{Average 1/5-shot 5-way few-shot classification accuracy over test images from the novel classes of the CIFAR-FS dataset. $^\ast$  indicate methods that train on a union of train and validation (train+val) set.
$^\dagger$: results from~\cite{bertinetto2018metalearning}.
$^\ddagger$: results from~\cite{gidaris2019boosting}.
}
\scalebox{0.75}{
\addtolength{\tabcolsep}{-4.5pt}
\begin{tabular}{|l|l|c|c|}
\hline
\multicolumn{1}{|l|}{Models} & Backbone & \multicolumn{1}{c|}{1-shot} & \multicolumn{1}{c|}{5-shot}\\

\hline
\hline
Proto Net~\cite{snell2017prototypical}$^\dagger$ (NIPS'17)& Conv-4-64  & 55.50 $\pm$ 0.70\% & 72.00 $\pm$ 0.60\%\\
Proto Net~\cite{snell2017prototypical}$^\dagger$ (NIPS'17)& Conv-4-512 & 57.90 $\pm$ 0.80\% & 76.70 $\pm$ 0.60\%\\
Proto Net~\cite{snell2017prototypical}$^\ddagger$ (NIPS'17) & Conv-4-64  & 62.82 $\pm$ 0.32\% & 79.59 $\pm$ 0.24\%\\
Proto Net~\cite{snell2017prototypical}$^\ddagger$    (NIPS'17)                 & Conv-4-512 & 66.48 $\pm$ 0.32\% & 80.28 $\pm$ 0.23\%\\
MAML~\cite{finn2017model}$^\dagger$   (ICML'17)                    & Conv-4-64  & 58.90 $\pm$ 1.90\% & 71.50 $\pm$ 1.00\%\\
MAML~\cite{finn2017model}$^\dagger$   (ICML'17)                    & Conv-4-512 & 53.80 $\pm$ 1.80\% & 67.60 $\pm$ 1.00\%\\
RelationNet~\cite{sung2018learning}$^\dagger$(CVPR'18)            & Conv-4-64  & 55.00 $\pm$ 1.00\% & 69.30 $\pm$ 0.80\%\\
GNN~\cite{garcia2018fewshot}$^\dagger$    (ICLR'18)                    & Conv-4-64  & 61.90\%            & 75.30\%\\
GNN~\cite{garcia2018fewshot}$^\dagger$    (ICLR'18)                    & Conv-4-512 & 56.00\%            & 72.50\%\\
R2-D2~\cite{bertinetto2018metalearning}     (ICLR'19)                          & Conv-4-64  & 62.30 $\pm$ 0.20\% & 77.40 $\pm$ 0.20\%\\
R2-D2~\cite{bertinetto2018metalearning}    (ICLR'19)                           & Conv-4-512 & 65.40 $\pm$ 0.20\% & 79.40 $\pm$ 0.20\%\\
Shot-Free~\cite{Ravichandran_2019} (ICCV'19) & ResNet-12 & 69.15\% & 84.70\%\\
MetaOptNet~\cite{lee2019meta}(CVPR'19)                & ResNet-12      & 72.60 $\pm$ 0.70\% & 84.30 $\pm$ 0.50\%\\
MetaOptNet$^\ast$~\cite{lee2019meta}(CVPR'19)                & ResNet-12      & 72.80 $\pm$ 0.70\% & 85.00 $\pm$ 0.50\%\\
BF3S CC+Rot~\cite{gidaris2019boosting}(ICCV'19)               & WRN-28-10  & 73.62 $\pm$ 0.31\% & 86.05 $\pm$ 0.22\%\\
RFS\cite{tian2020rethinking} & ResNet-12  & 73.89  $\pm$ 0.80\% & 86.93  $\pm$ 0.50\%\\
RFS\cite{tian2020rethinking}$^\ast$ & ResNet-12  & 75.40  $\pm$ 0.80\% & 88.20 $\pm$ 0.50\%\\
DSN-MR\cite{simon2020adaptive} (CVPR'20) & ResNet-12  & 75.60 $\pm$ 0.90\% & 86.20 $\pm$ 0.60\%\\
DSN-MR\cite{simon2020adaptive}$^\ast$ (CVPR'20) & ResNet-12  & 78.00 $\pm$ 0.90\% & 87.30 $\pm$ 0.60\%\\
\hline
\hline
BF3S CC\cite{gidaris2019boosting}+CRAT(Ours) & WRN-28-10  & 77.79 $\pm$ 0.32\% & 88.86 $\pm$ 0.23\%\\
\textbf{RFS\cite{tian2020rethinking} + CRAT} \textbf{(Ours)} & ResNet-12  & \textbf{77.18 $\pm$ 0.38\%} & \textbf{89.36 $\pm$ 0.25\%}\\
\textbf{RFS\cite{tian2020rethinking} + CRAT}$^\ast$ \textbf{(Ours)}  & ResNet-12  & \textbf{78.71 $\pm$ 0.35\%} & \textbf{89.78 $\pm$ 0.26\%}\\
\hline
\end{tabular}
}
\label{tab:cifarfs}
\vspace{-5pt}
\end{table}

 Few-shot classification performance on the CIFAR-FS dataset are shown in Table \ref{tab:cifarfs}. The results indicate that CC + CRAT performs better than CC + Rot by absolute margins of 4.17\% and 2.81\% for the 1-shot 5-way and 5-shot 5-way settings, respectively. RFS + CRAT outperforms RFS by absolute margins of 3.29\% and 2.43\% for the 1-shot 5-way and 5-shot 5-way settings. RFS + CRAT$^\ast$, with the model trained on train+val, performs better than RFS$^\ast$ (train+val) by absolute margins of 3.31\% and 1.58\% for the 1-shot 5-way and 5-shot 5-way settings respectively. The results also indicate that RFS + CRAT achieves state-of-the-art results.

\subsection{FC-100 Results}

\begin{table}[t]
\centering
\caption{Average 1/5-shot 5-way few-shot classification accuracy over test images from the novel classes of the FC-100 dataset. $^\ast$  indicate methods that train on a union of train and validation (train+val) set.}
\scalebox{0.75}{
\addtolength{\tabcolsep}{-4.5pt}
\begin{tabular}{|l|l|c|c|}
\hline
\multicolumn{1}{|l|}{Models} & Backbone & \multicolumn{1}{c|}{1-shot} & \multicolumn{1}{c|}{5-shot}\\

\hline
\hline
Proto Net~\cite{snell2017prototypical} (NIPS'17) & Conv-4-64  & 35.30 $\pm$ 0.60\% & 48.60 $\pm$ 0.60\%\\
TADAM~\cite{oreshkin2018tadam} (NIPS'18)                & ResNet-12      & 40.10 $\pm$ 0.40\% & 56.10 $\pm$ 0.40\%\\
MetaOptNet~\cite{lee2019meta}(CVPR'19)                & ResNet-12      & 41.10 $\pm$ 0.60\% & 55.50 $\pm$ 0.60\%\\
MetaOptNet$^\ast$~\cite{lee2019meta}(CVPR'19)                & ResNet-12      & 47.20 $\pm$ 0.60\% & 62.50 $\pm$ 0.60\%\\
MTL~\cite{Sun_2019} (CVPR'19)                & ResNet-12      & 45.10 $\pm$ 1.80\% & 57.60 $\pm$ 0.90\%\\
DC~\cite{Lifchitz_2019} (CVPR'19)                & ResNet-12      & 42.04 $\pm$ 0.17\% & 57.63 $\pm$ 0.23\%\\
RFS\cite{tian2020rethinking} & ResNet-12  & 44.57  $\pm$ 0.70\% & 60.91  $\pm$ 0.60\%\\
RFS\cite{tian2020rethinking}$^\ast$ & ResNet-12  & 51.60  $\pm$ 0.70\% & 68.40 $\pm$ 0.60\%\\
Transductive\cite{Dhillon2020A} (ICLR'20) & WRN-28-10  & 43.16 $\pm$ 0.59\% &  57.57 $\pm$ 0.55\%\\
Transductive\cite{Dhillon2020A}$^\ast$ (ICLR'20) & WRN-28-10  & 50.44 $\pm$ 0.68\% &  65.74 $\pm$ 0.60\%\\
\hline
\hline
\textbf{RFS\cite{tian2020rethinking} + CRAT} \textbf{(Ours)} & ResNet-12  & \textbf{46.85 $\pm$ 0.35\%} & \textbf{63.56 $\pm$ 0.33\%}\\
\textbf{RFS\cite{tian2020rethinking} + CRAT}$^\ast$ \textbf{(Ours)}  & ResNet-12  & \textbf{54.70 $\pm$ 0.38\%} & \textbf{70.76 $\pm$ 0.35\%}\\
\hline
\end{tabular}
}
\label{tab:fc100}
\vspace{-15pt}
\end{table}

Table \ref{tab:fc100} depicts the results for few-shot classification on the FC-100 dataset. From the results, it can be seen that RFS + CRAT performs better than RFS by absolute margins of 2.28\% and 2.65\% for the 1-shot 5-way and 5-shot 5-way settings, respectively. RFS + CRAT$^\ast$, with the model trained on train+val, performs better than RFS$^\ast$ (train+val) by absolute margins of 3.1\% and 2.36\% for the 1-shot 5-way and 5-shot 5-way settings respectively. It also performs better than existing state-of-the-art methods by a significant margin.

\begin{figure*}[t]
    \centering
    \includegraphics[scale=.45]{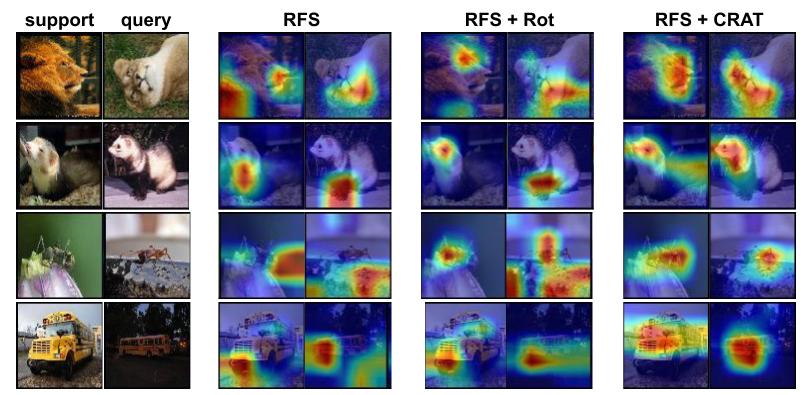} 
    \caption{Class activation mapping visualization for novel class images of the mini-ImageNet dataset, using the feature extraction network of RFS, RFS + Rot (auxiliary rotation task), and RFS + CRAT (auxiliary composite rotation task) with ResNet-12 architecture.}
\label{fig:qual}
\vspace{-10pt}
\end{figure*}

\subsection{Ablations}
We perform ablation experiments: 1) to validate the transformations involved in our composite rotation operation 2) to compare our composite rotation to other self-supervised auxiliary tasks.
\vspace{-5pt}

\subsubsection{Transformations in Composite Rotation}\label{sec:ablation}

Our proposed composite rotation involves rotating the entire image by either 0, 90, 180, or 270 degrees, splitting the image vertically from the middle, and finally, rotating each half by either 0 or 180 degrees (see Fig. \ref{fig:method}). We perform ablations to validate our transformation choices by using various combinations of transformations as the auxiliary task used along with RFS on the CIFAR-FS dataset with ResNet-12 architecture. RFS + Rot uses the simple rotation based self-supervision as used in \cite{gidaris2019boosting} and has 4 classes. RFS + HS4 splits the image horizontally from the middle and rotates each half by either 0 or 180 degrees, resulting in 4 classes. Similarly, RFS + VS4 splits the image vertically from the middle and rotates each half by either 0 or 180 degrees. RFS + HS16 and RFS + VS16 rotate the entire image by either 0, 90, 180, or 270 degrees apart from splitting the image horizontally and vertically, respectively, and rotating each half by 0 or 180 degrees. This results in 16 classes. RFS + Rot32 combines RFS + HS16 and RFS + VS16 classes to obtain 32 classes of transformation. RFS + Rot256 splits the image into 4 quarters and rotates each quarter by either 0, 90, 180, or 270 degrees resulting in 256 classes. Similarly, RFS + Rot1024 splits the image into 4 quarters and rotates each quarter by either 0, 90, 180, or 270 degrees and also rotates the entire image by either 0, 90, 180, or 270 degrees resulting in 1024 classes. 

\begin{table}[t]
\centering
\caption{Average 1/5-shot 5-way few-shot classification accuracy over test images from the novel classes of the CIFAR-FS for auxiliary tasks based on different transformations. Experiments are conducted using RFS and ResNet-12 architecture.}
\scalebox{0.75}{
\addtolength{\tabcolsep}{-1.5pt}
\begin{tabular}{|l|c|c|c|}
\hline
\multicolumn{1}{|l|}{Models} & \multicolumn{1}{|l|}{Classes} & \multicolumn{1}{c|}{1-shot} & \multicolumn{1}{c|}{5-shot}\\
\hline
\hline
RFS + Rot & 4 & 74.71 $\pm$ 0.69\% & 87.92 $\pm$ 0.26\% \\
RFS + HS4 & 4 & 74.12 $\pm$ 0.66\% & 87.20 $\pm$ 0.27\%\\
RFS + VS4 & 4 & 74.45 $\pm$ 0.65\% & 87.50 $\pm$ 0.25\%\\
RFS + HS16 & 16 &  77.02 $\pm$ 0.35\% & 89.15 $\pm$ 0.26\%\\
RFS + VS16 (CRAT) & 16 &  77.18 $\pm$ 0.38\% & 89.36 $\pm$ 0.25\%\\
RFS + Rot32 (HS16 + VS16) & 32 & 77.31 $\pm$ 0.41\% & 89.51 $\pm$ 0.29\%\\
RFS + Rot256 & 256 & 73.56 $\pm$ 0.40\% & 86.05 $\pm$ 0.28\%\\
RFS + Rot1024 & 1024 & 73.77 $\pm$ 0.72\% & 86.36 $\pm$ 0.45\%\\
\hline
\end{tabular}}

\label{tab:ablation_transformation}
\vspace{-15pt}
\end{table}

Table \ref{tab:ablation_transformation} indicates that RFS + HS4 and RFS + VS4 perform close to RFS + Rot. Therefore, splitting an image into half and rotating each half performs similar to the simple rotation operation, when used as an auxiliary task. RFS + HS16 and RFS + VS16 perform significantly better than RFS + Rot. RFS + VS16 performs slightly better than RFS + HS16, and we choose VS16 for our composite rotation operation. 
In RFS + Rot32, each image is converted to 32 different transformed samples. Since the performance of RFS + Rot32 is similar to RFS + VS16 and the memory overhead is high for RFS + Rot32, we use RFS + VS16 in our proposed approach. RFS + Rot256 and RFS + Rot1024 convert each image into 256 and 1024 different transformed samples respectively. This requires huge computing resources, which is not always feasible. In order to implement them in a scalable way, we randomly apply only 1 transformation class out of 256 in the case of Rot256 and 1 out of 1024 in the case of Rot1024 to each image. The results indicate that Rot1024 and Rot256 are unable to provide improvements to the base model. This possibly results from the large number of classes in the rotation classifier, which makes the auxiliary task very complicated and difficult and consequently hurts the classifier's performance since they share the same feature extraction network.

\vspace{-5pt}

\subsubsection{Comparison with Other Self-Supervised Auxiliary Tasks}\label{sec:ablation_ss}
\begin{table}[t]
\centering
\caption{Average 1/5-shot 5-way few-shot classification accuracy over test images from the novel classes of the CIFAR-FS for different types of auxiliary self-supervised tasks on RFS with ResNet-12 architecture.}
\scalebox{0.75}{
\addtolength{\tabcolsep}{5pt}
\begin{tabular}{|l|c|c|}
\hline
\multicolumn{1}{|l|}{Models} & \multicolumn{1}{c|}{1-shot} & \multicolumn{1}{c|}{5-shot}\\
\hline
\hline
RFS + Patch \cite{doersch2015unsupervised} &  74.23 $\pm$ 0.65\% & 87.33 $\pm$ 0.35\%\\
RFS + Rot \cite{gidaris2019boosting,gidaris2018unsupervised}&  74.71 $\pm$ 0.69\% & 87.92 $\pm$ 0.26\% \\
RFS + SimCLR \cite{chen2020simple}&  74.45 $\pm$ 0.67\% & 87.42 $\pm$ 0.30\%\\
\textbf{RFS + CRAT (Ours)} &  \textbf{77.18 $\pm$ 0.38\%} & \textbf{89.36 $\pm$ 0.25\%}\\  
\hline
\end{tabular}}

\label{tab:ablation_ss}
\vspace{-20pt}
\end{table}

We compare our proposed composite rotation based auxiliary task (CRAT) with other auxiliary tasks based on self-supervised techniques. We perform experiments on RFS \cite{tian2020rethinking} with the auxiliary task as relative patch location prediction \cite{doersch2015unsupervised} (RFS + Patch), simple rotation (RFS + Rot) \cite{gidaris2019boosting,gidaris2018unsupervised}  and SimCLR \cite{chen2020simple}. SimCLR is the current state-of-the-art self-supervised training technique that uses contrastive learning. SimCLR can be easily used as an auxiliary task, which is a necessary pre-condition for our approach.

Table \ref{tab:ablation_ss}, shows that RFS + CRAT performs significantly better than RFS + Patch, RFS + Rot and RFS + SimCLR. Even though SimCLR is a state-of-the-art self-supervision method, it is unable to outperform CRAT when used as an auxiliary task for few-shot learning.

\subsection{Qualitative Results}
Fig.~\ref{fig:qual}, shows the comparison of the class activation map \cite{selvaraju2017grad} visualizations for RFS \cite{tian2020rethinking}, RFS with simple rotation based auxiliary task (RFS + Rot) and RFS with our proposed composite rotation based auxiliary task (RFS + CRAT). The class activation mappings are visualized for the novel class images of the mini-ImageNet dataset, and the networks have not been trained on these classes. The visualizations show that RFS + CRAT attends more to the discriminative regions of the object, and this helps the network in extracting more discriminative and generic features, which in turn helps to improve the few-shot classification performance.

\section{Conclusion}
We propose a technique of improving few-shot classification by using a composite rotation based auxiliary task (CRAT). Our approach involves training the feature extraction network on this auxiliary task along with the general classification task. We demonstrate the efficiency of CRAT by plugging it into two recent few-shot learning methods \cite{tian2020rethinking,gidaris2019boosting}. We experimentally show that our proposed auxiliary task significantly improves both of these few-shot learning methods. We show that RFS + CRAT outperforms existing state-of-the-art few-shot learning methods through several experiments on multiple benchmark datasets. We also validate our proposed composite rotation by performing ablation experiments.

{\small
\bibliographystyle{ieee_fullname}
\bibliography{egbib}

\begin{thebibliography}{10}\itemsep=-1pt

\bibitem{bertinetto2018metalearning}
Luca Bertinetto, Joao~F. Henriques, Philip Torr, and Andrea Vedaldi.
\newblock Meta-learning with differentiable closed-form solvers.
\newblock In {\em International Conference on Learning Representations}, 2019.

\bibitem{chen2019variational}
Jiaxin Chen, Li-Ming Zhan, Xiao-Ming Wu, and Fu lai Chung.
\newblock Variational metric scaling for metric-based meta-learning.
\newblock In {\em Proceedings of the AAAI Conference on Artificial
  Intelligence}, 2020.

\bibitem{chen2019diversity}
Mengting Chen, Yuxin Fang, Xinggang Wang, Heng Luo, Yifeng Geng, Xinyu Zhang,
  Chang Huang, Wenyu Liu, and Bo Wang.
\newblock Diversity transfer network for few-shot learning.
\newblock In {\em Proceedings of the AAAI Conference on Artificial
  Intelligence}, 2020.

\bibitem{chen2020simple}
Ting Chen, Simon Kornblith, Mohammad Norouzi, and Geoffrey Hinton.
\newblock A simple framework for contrastive learning of visual
  representations.
\newblock {\em arXiv preprint arXiv:2002.05709}, 2020.

\bibitem{Chen_2019}
Zitian Chen, Yanwei Fu, Yu-Xiong Wang, Lin Ma, Wei Liu, and Martial Hebert.
\newblock Image deformation meta-networks for one-shot learning.
\newblock In {\em Proceedings of the IEEE Conference on Computer Vision and
  Pattern Recognition (CVPR)}, pages 8680--8689, 2019.

\bibitem{Dhillon2020A}
Guneet~Singh Dhillon, Pratik Chaudhari, Avinash Ravichandran, and Stefano
  Soatto.
\newblock A baseline for few-shot image classification.
\newblock In {\em International Conference on Learning Representations}, 2020.

\bibitem{doersch2015unsupervised}
Carl Doersch, Abhinav Gupta, and Alexei~A Efros.
\newblock Unsupervised visual representation learning by context prediction.
\newblock In {\em Proceedings of the IEEE International Conference on Computer
  Vision}, pages 1422--1430, 2015.

\bibitem{dosovitskiy2014discriminative}
Alexey Dosovitskiy, Jost~Tobias Springenberg, Martin Riedmiller, and Thomas
  Brox.
\newblock Discriminative unsupervised feature learning with convolutional
  neural networks.
\newblock In {\em Advances in neural information processing systems}, pages
  766--774, 2014.

\bibitem{feng2019self}
Zeyu Feng, Chang Xu, and Dacheng Tao.
\newblock Self-supervised representation learning by rotation feature
  decoupling.
\newblock In {\em Proceedings of the IEEE Conference on Computer Vision and
  Pattern Recognition}, pages 10364--10374, 2019.

\bibitem{finn2017model}
Chelsea Finn, Pieter Abbeel, and Sergey Levine.
\newblock Model-agnostic meta-learning for fast adaptation of deep networks.
\newblock In {\em Proceedings of the 34th International Conference on Machine
  Learning-Volume 70}, pages 1126--1135. JMLR. org, 2017.

\bibitem{Flennerhag2020Meta-Learning}
Sebastian Flennerhag, Andrei~A. Rusu, Razvan Pascanu, Francesco Visin, Hujun
  Yin, and Raia Hadsell.
\newblock Meta-learning with warped gradient descent.
\newblock In {\em International Conference on Learning Representations}, 2020.

\bibitem{gidaris2019boosting}
Spyros Gidaris, Andrei Bursuc, Nikos Komodakis, Patrick P{\'e}rez, and Matthieu
  Cord.
\newblock Boosting few-shot visual learning with self-supervision.
\newblock In {\em Proceedings of the IEEE International Conference on Computer
  Vision}, pages 8059--8068, 2019.

\bibitem{gidaris2018dynamic}
Spyros Gidaris and Nikos Komodakis.
\newblock Dynamic few-shot visual learning without forgetting.
\newblock In {\em Proceedings of the IEEE Conference on Computer Vision and
  Pattern Recognition}, pages 4367--4375, 2018.

\bibitem{gidaris2018unsupervised}
Spyros Gidaris, Praveer Singh, and Nikos Komodakis.
\newblock Unsupervised representation learning by predicting image rotations.
\newblock In {\em International Conference on Learning Representations}, 2018.

\bibitem{guo2020attentive}
Yiluan Guo and Ngai-Man Cheung.
\newblock Attentive weights generation for few shot learning via information
  maximization.
\newblock In {\em Proceedings of the IEEE/CVF Conference on Computer Vision and
  Pattern Recognition}, pages 13499--13508, 2020.

\bibitem{he2020momentum}
Kaiming He, Haoqi Fan, Yuxin Wu, Saining Xie, and Ross Girshick.
\newblock Momentum contrast for unsupervised visual representation learning.
\newblock In {\em Proceedings of the IEEE/CVF Conference on Computer Vision and
  Pattern Recognition}, pages 9729--9738, 2020.

\bibitem{krizhevsky2009learning}
Alex Krizhevsky, Geoffrey Hinton, et~al.
\newblock Learning multiple layers of features from tiny images.
\newblock 2009.

\bibitem{krizhevsky2012imagenet}
Alex Krizhevsky, Ilya Sutskever, and Geoffrey~E Hinton.
\newblock Imagenet classification with deep convolutional neural networks.
\newblock In {\em Advances in neural information processing systems}, pages
  1097--1105, 2012.

\bibitem{larsson2016learning}
Gustav Larsson, Michael Maire, and Gregory Shakhnarovich.
\newblock Learning representations for automatic colorization.
\newblock In {\em European Conference on Computer Vision}, pages 577--593.
  Springer, 2016.

\bibitem{lee2019meta}
Kwonjoon Lee, Subhransu Maji, Avinash Ravichandran, and Stefano Soatto.
\newblock Meta-learning with differentiable convex optimization.
\newblock In {\em Proceedings of the IEEE Conference on Computer Vision and
  Pattern Recognition (CVPR)}, pages 10657--10665, 2019.

\bibitem{li2020adversarial}
Kai Li, Yulun Zhang, Kunpeng Li, and Yun Fu.
\newblock Adversarial feature hallucination networks for few-shot learning.
\newblock In {\em Proceedings of the IEEE/CVF Conference on Computer Vision and
  Pattern Recognition}, pages 13470--13479, 2020.

\bibitem{Lifchitz_2019}
Yann Lifchitz, Yannis Avrithis, Sylvaine Picard, and Andrei Bursuc.
\newblock Dense classification and implanting for few-shot learning.
\newblock In {\em Proceedings of the IEEE Conference on Computer Vision and
  Pattern Recognition (CVPR)}, pages 9258--9267, 2019.

\bibitem{liu2019learning}
Yanbin Liu, Juho Lee, Minseop Park, Saehoon Kim, Eunho Yang, Sungju Hwang, and
  Yi Yang.
\newblock Learning to propagate labels: Transductive propagation network for
  few-shot learning.
\newblock In {\em International Conference on Learning Representations}, 2019.

\bibitem{mishra2018a}
Nikhil Mishra, Mostafa Rohaninejad, Xi Chen, and Pieter Abbeel.
\newblock A simple neural attentive meta-learner.
\newblock In {\em International Conference on Learning Representations}, 2018.

\bibitem{munkhdalai2017meta}
Tsendsuren Munkhdalai and Hong Yu.
\newblock Meta networks.
\newblock In {\em Proceedings of the 34th International Conference on Machine
  Learning-Volume 70}, pages 2554--2563. JMLR. org, 2017.

\bibitem{noroozi2016unsupervised}
Mehdi Noroozi and Paolo Favaro.
\newblock Unsupervised learning of visual representations by solving jigsaw
  puzzles.
\newblock In {\em European Conference on Computer Vision}, pages 69--84.
  Springer, 2016.

\bibitem{oreshkin2018tadam}
Boris Oreshkin, Pau~Rodr{\'\i}guez L{\'o}pez, and Alexandre Lacoste.
\newblock Tadam: Task dependent adaptive metric for improved few-shot learning.
\newblock In {\em Advances in Neural Information Processing Systems}, pages
  721--731, 2018.

\bibitem{pathak2016context}
Deepak Pathak, Philipp Krahenbuhl, Jeff Donahue, Trevor Darrell, and Alexei~A
  Efros.
\newblock Context encoders: Feature learning by inpainting.
\newblock In {\em Proceedings of the IEEE conference on computer vision and
  pattern recognition}, pages 2536--2544, 2016.

\bibitem{qiao2018few}
Siyuan Qiao, Chenxi Liu, Wei Shen, and Alan~L Yuille.
\newblock Few-shot image recognition by predicting parameters from activations.
\newblock In {\em Proceedings of the IEEE Conference on Computer Vision and
  Pattern Recognition}, pages 7229--7238, 2018.

\bibitem{Ravichandran_2019}
Avinash Ravichandran, Rahul Bhotika, and Stefano Soatto.
\newblock Few-shot learning with embedded class models and shot-free meta
  training.
\newblock In {\em Proceedings of the IEEE International Conference on Computer
  Vision (ICCV)}, pages 331--339, 2019.

\bibitem{ren2018metalearning}
Mengye Ren, Sachin Ravi, Eleni Triantafillou, Jake Snell, Kevin Swersky,
  Josh~B. Tenenbaum, Hugo Larochelle, and Richard~S. Zemel.
\newblock Meta-learning for semi-supervised few-shot classification.
\newblock In {\em International Conference on Learning Representations}, 2018.

\bibitem{russakovsky2015imagenet}
Olga Russakovsky, Jia Deng, Hao Su, Jonathan Krause, Sanjeev Satheesh, Sean Ma,
  Zhiheng Huang, Andrej Karpathy, Aditya Khosla, Michael Bernstein, et~al.
\newblock Imagenet large scale visual recognition challenge.
\newblock {\em International journal of computer vision}, 115(3):211--252,
  2015.

\bibitem{rusu2018metalearning}
Andrei~A. Rusu, Dushyant Rao, Jakub Sygnowski, Oriol Vinyals, Razvan Pascanu,
  Simon Osindero, and Raia Hadsell.
\newblock Meta-learning with latent embedding optimization.
\newblock In {\em International Conference on Learning Representations}, 2019.

\bibitem{garcia2018fewshot}
Victor~Garcia Satorras and Joan~Bruna Estrach.
\newblock Few-shot learning with graph neural networks.
\newblock In {\em International Conference on Learning Representations}, 2018.

\bibitem{selvaraju2017grad}
Ramprasaath~R Selvaraju, Michael Cogswell, Abhishek Das, Ramakrishna Vedantam,
  Devi Parikh, and Dhruv Batra.
\newblock Grad-cam: Visual explanations from deep networks via gradient-based
  localization.
\newblock In {\em Proceedings of the IEEE international conference on computer
  vision}, pages 618--626, 2017.

\bibitem{simon2020adaptive}
Christian Simon, Piotr Koniusz, Richard Nock, and Mehrtash Harandi.
\newblock Adaptive subspaces for few-shot learning.
\newblock In {\em Proceedings of the IEEE/CVF Conference on Computer Vision and
  Pattern Recognition}, pages 4136--4145, 2020.

\bibitem{snell2017prototypical}
Jake Snell, Kevin Swersky, and Richard Zemel.
\newblock Prototypical networks for few-shot learning.
\newblock In {\em Advances in Neural Information Processing Systems}, pages
  4077--4087, 2017.

\bibitem{Sun_2019}
Qianru Sun, Yaoyao Liu, Tat-Seng Chua, and Bernt Schiele.
\newblock Meta-transfer learning for few-shot learning.
\newblock In {\em Proceedings of the IEEE Conference on Computer Vision and
  Pattern Recognition (CVPR)}, pages 403--412, 2019.

\bibitem{sung2018learning}
Flood Sung, Yongxin Yang, Li Zhang, Tao Xiang, Philip~HS Torr, and Timothy~M
  Hospedales.
\newblock Learning to compare: Relation network for few-shot learning.
\newblock In {\em Proceedings of the IEEE Conference on Computer Vision and
  Pattern Recognition}, pages 1199--1208, 2018.

\bibitem{tian2019contrastive}
Yonglong Tian, Dilip Krishnan, and Phillip Isola.
\newblock Contrastive multiview coding.
\newblock {\em arXiv preprint arXiv:1906.05849}, 2019.

\bibitem{tian2020rethinking}
Yonglong Tian, Yue Wang, Dilip Krishnan, Joshua~B. Tenenbaum, and Phillip
  Isola.
\newblock Rethinking few-shot image classification: a good embedding is all you
  need?
\newblock {\em arXiv preprint arXiv:2003.11539}, 2020.

\bibitem{vinyals2016matching}
Oriol Vinyals, Charles Blundell, Timothy Lillicrap, Daan Wierstra, et~al.
\newblock Matching networks for one shot learning.
\newblock In {\em Advances in neural information processing systems}, pages
  3630--3638, 2016.

\bibitem{Yan2019ADA}
Shipeng Yan, Songyang Zhang, Xuming He, et~al.
\newblock A dual attention network with semantic embedding for few-shot
  learning.
\newblock In {\em Proceedings of the AAAI Conference on Artificial
  Intelligence}, volume~33, pages 9079--9086, 2019.

\bibitem{zagoruyko2016wide}
Sergey Zagoruyko and Nikos Komodakis.
\newblock Wide residual networks.
\newblock In Edwin R.~Hancock Richard C.~Wilson and William A.~P. Smith,
  editors, {\em Proceedings of the British Machine Vision Conference (BMVC)},
  pages 87.1--87.12. BMVA Press, September 2016.

\bibitem{Zhang_2019}
Hongguang Zhang, Jing Zhang, and Piotr Koniusz.
\newblock Few-shot learning via saliency-guided hallucination of samples.
\newblock In {\em Proceedings of the IEEE Conference on Computer Vision and
  Pattern Recognition (CVPR)}, pages 2770--2779, 2019.

\bibitem{zhang2016colorful}
Richard Zhang, Phillip Isola, and Alexei~A Efros.
\newblock Colorful image colorization.
\newblock In {\em European conference on computer vision}, pages 649--666.
  Springer, 2016.

\bibitem{zhou2014learning}
Bolei Zhou, Agata Lapedriza, Jianxiong Xiao, Antonio Torralba, and Aude Oliva.
\newblock Learning deep features for scene recognition using places database.
\newblock In {\em Advances in neural information processing systems}, pages
  487--495, 2014.

\end{thebibliography}
}

\end{document}